\documentclass[letterpaper]{article} 
\usepackage[T1]{fontenc}    
\usepackage[utf8]{inputenc}

\usepackage{aaai2026}
\usepackage{times}  
\usepackage{helvet}  
\usepackage{courier}  
\usepackage[hyphens]{url}  
\usepackage{graphicx} 
\usepackage{natbib}  
\usepackage{caption} 
\usepackage{algorithm}
\usepackage{algorithmic}
\usepackage{xcolor}         
\usepackage{graphicx}
\usepackage{amsmath}
\usepackage{multirow}
\usepackage{booktabs}

\usepackage{times}
\usepackage{helvet}
\usepackage{courier}
\usepackage{xcolor}
\usepackage{newfloat}
\usepackage{listings}
\DeclareCaptionStyle{ruled}{labelfont=normalfont,labelsep=colon,strut=off} 
\floatstyle{ruled}
\newfloat{listing}{tb}{lst}{}
\floatname{listing}{Listing}
\title{Undress to Redress: A Training-Free Framework for Virtual Try-On}
\author{
    Zhiying Li\textsuperscript{\rm 1}, Junhao Wu\textsuperscript{\rm 3}, Yeying Jin\textsuperscript{\rm 4}, Daiheng Gao\textsuperscript{\rm 5}, Yun Ji\textsuperscript{\rm 2}, Kaichuan Kong\textsuperscript{\rm 1},\\
    Lei Yu\textsuperscript{\rm 6}, Hao Xu\textsuperscript{\rm 7}, Kai Chen\textsuperscript{\rm 7}, Bruce Gu\textsuperscript{\rm 8}, Nana Wang\textsuperscript{\rm 9}, Zhaoxin Fan\textsuperscript{\rm 2}
}
\affiliations{
    \textsuperscript{\rm 1}Jinan University; \textsuperscript{\rm 2}Beihang University; \textsuperscript{\rm 3}Psyche AI Inc; \textsuperscript{\rm 4}Tencent;\\
    \textsuperscript{\rm 5}University of Science and Technology of China; \textsuperscript{\rm 6}Anhui University; \\
    \textsuperscript{\rm 7}Hong Kong University of Science and Technology; 
    \textsuperscript{\rm 8}Shandong Computer Science Center; \textsuperscript{\rm 9}Eliza Lab
    
}

\usepackage{bibentry}

\begin{document}

\maketitle

\begin{abstract}
Virtual try-on (VTON) is a crucial task for enhancing user experience in online shopping by generating realistic garment previews on personal photos. Although existing methods have achieved impressive results, they struggle with long-sleeve-to-short-sleeve conversions—a common and practical scenario—often producing unrealistic outputs when exposed skin is underrepresented in the original image. We argue that this challenge arises from the “majority” completion rule in current VTON models, which leads to inaccurate skin restoration in such cases. To address this, we propose UR-VTON (\textbf{U}ndress-\textbf{R}edress \textbf{V}irtual \textbf{T}ry-\textbf{ON}), a novel, training-free framework that can be seamlessly integrated with any existing VTON method. UR-VTON introduces an ``undress-to-redress'' mechanism: it first reveals the user's torso by virtually ``undressing,'' then applies the target short-sleeve garment, effectively decomposing the conversion into two more manageable steps. Additionally, we incorporate Dynamic Classifier-Free Guidance scheduling to balance diversity and image quality during DDPM sampling, and employ Structural Refiner to enhance detail fidelity using high-frequency cues. Finally, we present LS-TON, a new benchmark for long-sleeve-to-short-sleeve try-on. Extensive experiments demonstrate that UR-VTON outperforms state-of-the-art methods in both detail preservation and image quality. Code will be released upon acceptance.
\end{abstract}

\section{Introduction}
\begin{figure}[t]
    \centering
    \includegraphics[width=0.45\textwidth]{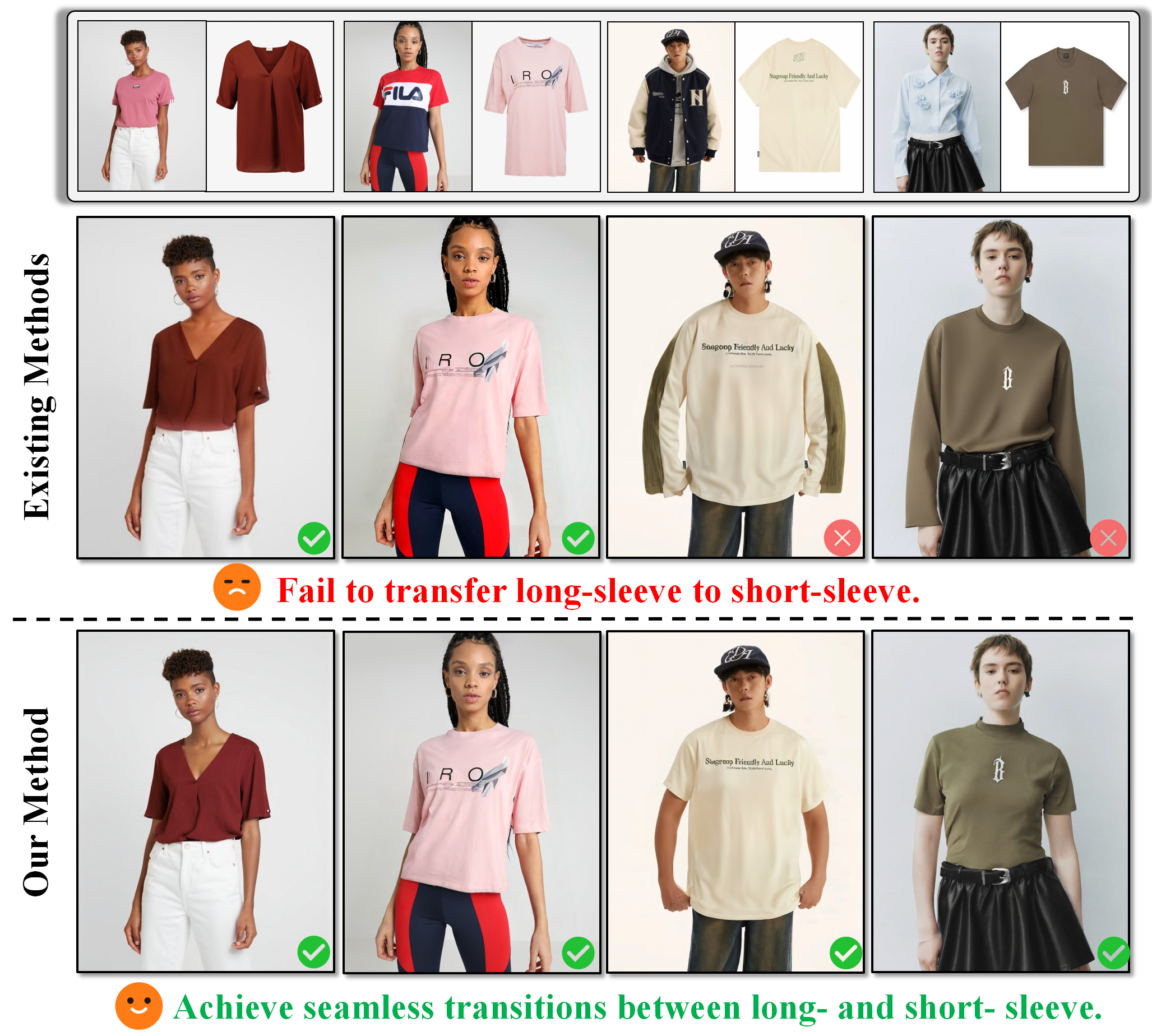}
    \caption{Illustration of virtual try-on.
Existing methods frequently fail to produce realistic results when transforming long-sleeve to short-sleeve. In contrast, our method achieves realistic virtual try-on with seamless transitions between long- and short- sleeve.}
    \label{motivation}
\end{figure}

Virtual try-on (VTON) realistically renders specified garments onto user-supplied images, significantly enhancing online shopping. However, existing methods focus on short-sleeve-to-short-sleeve or short-sleeve-to-long-sleeve transformations, common in summer and autumn, and perform poorly when users upload long-sleeve images yet seek to try on designated short sleeves during the spring–summer transition. These approaches often produce inaccurate garment replacements and introduce artifacts (see Fig. \ref{motivation}), severely limiting VTON’s practical utility and its ability to meet growing consumer demand for personalized, seamless shopping interactions. Thus, effective long-to-short sleeve conversion remains an urgent challenge.

Early VTON methods predominantly employed geometric deformation techniques, such as thin-plate splines (TPS) \cite{bookstein2002principal, ge2021disentangled, han2018viton} or optical-flow-based approaches \cite{bai2022single, ge2021parser, he2022style}, but they often suffer from misalignment and fail to produce truly realistic garment swaps. Recently, inspired by the advances in conditional generative methods \cite{rombach2022high, li2023blip}, a variety of new VTON approaches \cite{kim2024stableviton, zhu2023tryondiffusion, yang2024texture} have emerged and achieved substantial improvements in visual quality. For instance, MGD \cite{baldrati2023multimodal} introduce the first latent-space diffusion model for human fashion image editing, supporting multimodal conditioning through text, pose, and sketch inputs; LaDI-VTON \cite{morelli2023ladi} and StableVITON \cite{kim2024stableviton} adopt ControlNet \cite{zhang2023adding} architectures to encode supplementary information. DCI-VTON \cite{gou2023taming} casts virtual try-on as an image-inpainting task by incorporating the deformed garment as a local condition within the diffusion model. Similarly, OOTDiffusion  \cite{xu2025ootdiffusion}, IDM-VTON \cite{choi2024improving}, and OutfitAnyone \cite{sun2024outfitanyone} utilize reference networks akin to denoising U-Nets \cite{ronneberger2015u} to generate new clothing appearances by processing garment regions. Despite these significant advances, existing methods still struggle with the common and practical challenge of converting long sleeves to short sleeves. We attribute this difficulty to the ``majority'' completion rule in current VTON models: if most pixels to be filled correspond to skin, the model correctly restores skin regions; but when the majority of the target region is clothing, it often erroneously completes skin areas with garment textures (as illustrated in Fig. \ref{motivation}). This limitation restricts VTON’s applicability to general scenarios and hinders accurate long-sleeve to short-sleeve try-on.

To address this challenge, we propose Undress-Redress Virtual Try-On (UR-VTON), a novel, training-free framework that can be seamlessly integrated with any existing VTON model. At the core of our approach is an ``undress-to-redress'' mechanism: we first undress the long-sleeved garment to reveal the torso, and then redress it with the target short-sleeved garment. By decomposing the challenging one-step transformation from long sleeves to short sleeves into two manageable stages, our method substantially improves both accuracy and consistency. To further enhance performance, we introduce a dynamic classifier-free guidance scheduler, which addresses the limitations of using a fixed guidance scale in DDPM sampling \cite{ho2020denoising}—a common issue that often struggles to balance diversity and image quality \cite{malarz2025classifier}. Complementing this, we present a structural refiner strategy that leverages high-frequency structural cues from the input image at the final output stage, further enhancing visual fidelity. To facilitate comprehensive evaluation, we also introduce LS-TON, a novel benchmark specifically designed for long-sleeve-to-short-sleeve try-on scenarios. Extensive experiments demonstrate that UR-VTON achieves state-of-the-art results, yielding significant gains in detail preservation and overall image quality. In summary, our contribution can be summarized as: 
\begin{itemize}
    \item We propose UR-VTON, a novel, training-free framework that can be seamlessly integrated with any existing VTON model. By employing an undress-to-redress mechanism, it decomposes the complex, one-step task of converting long sleeves to short sleeves into simpler, controllable steps.
    
    \item We employ a dynamic classifier-free guidance schedule to balance diversity and image quality during DDPM sampling, and utilize structural refiner to enhance detail fidelity by leveraging high-frequency structural cues.
    
    \item We present LS-TON, a novel benchmark comprising 770 image triplets, each consisting of a model with long-sleeve, the corresponding short-sleeved garment, and the model with short-sleeve, for evaluating long-sleeve-to-short-sleeve try-on performance.
    
    \item Extensive experiments demonstrate that UR-VTON attains state-of-the-art performance, yielding substantial improvements in detail preservation and overall image quality, and validating its practical applicability.
\end{itemize}

\begin{figure*}[t]
    \centering
    \includegraphics[width=0.90\textwidth]{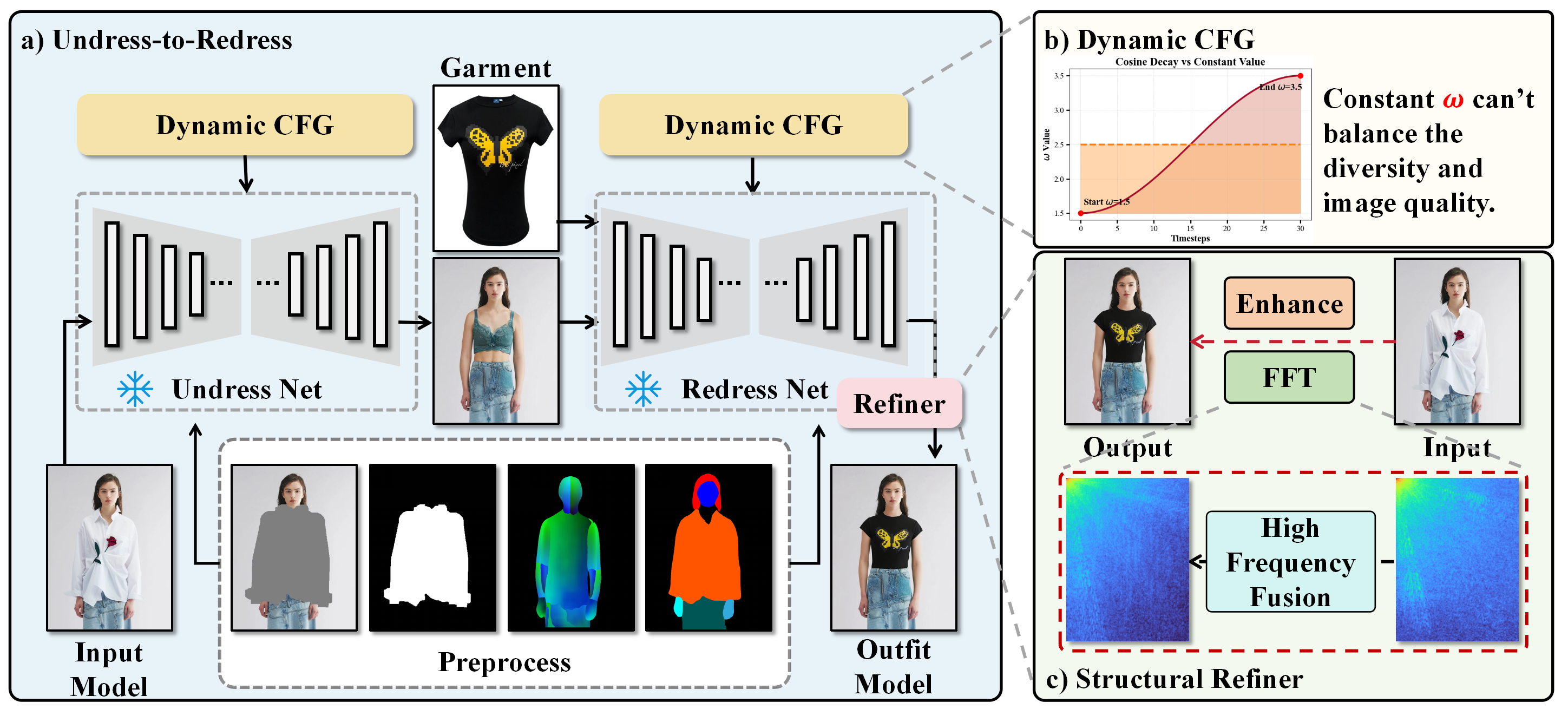}
    \caption{Overview of the proposed UR-VTON framework. \textbf{a)} UR-VTON employs an ``Undress-to-Redress'' mechanism: it first undresses the long-sleeved garment to expose the torso and then redresses the target short-sleeved garment. \textbf{b)} Dynamic CFG scheduler is incorporated into the diffusion process to balance diversity and image quality. \textbf{c)} Structural Refiner leverages high-frequency structural cues from the input image at the final output stage to further enhance visual fidelity.}
    \label{UR-VTON}
\end{figure*}

\begin{figure}[t]
    \centering
    \includegraphics[width=0.45\textwidth]{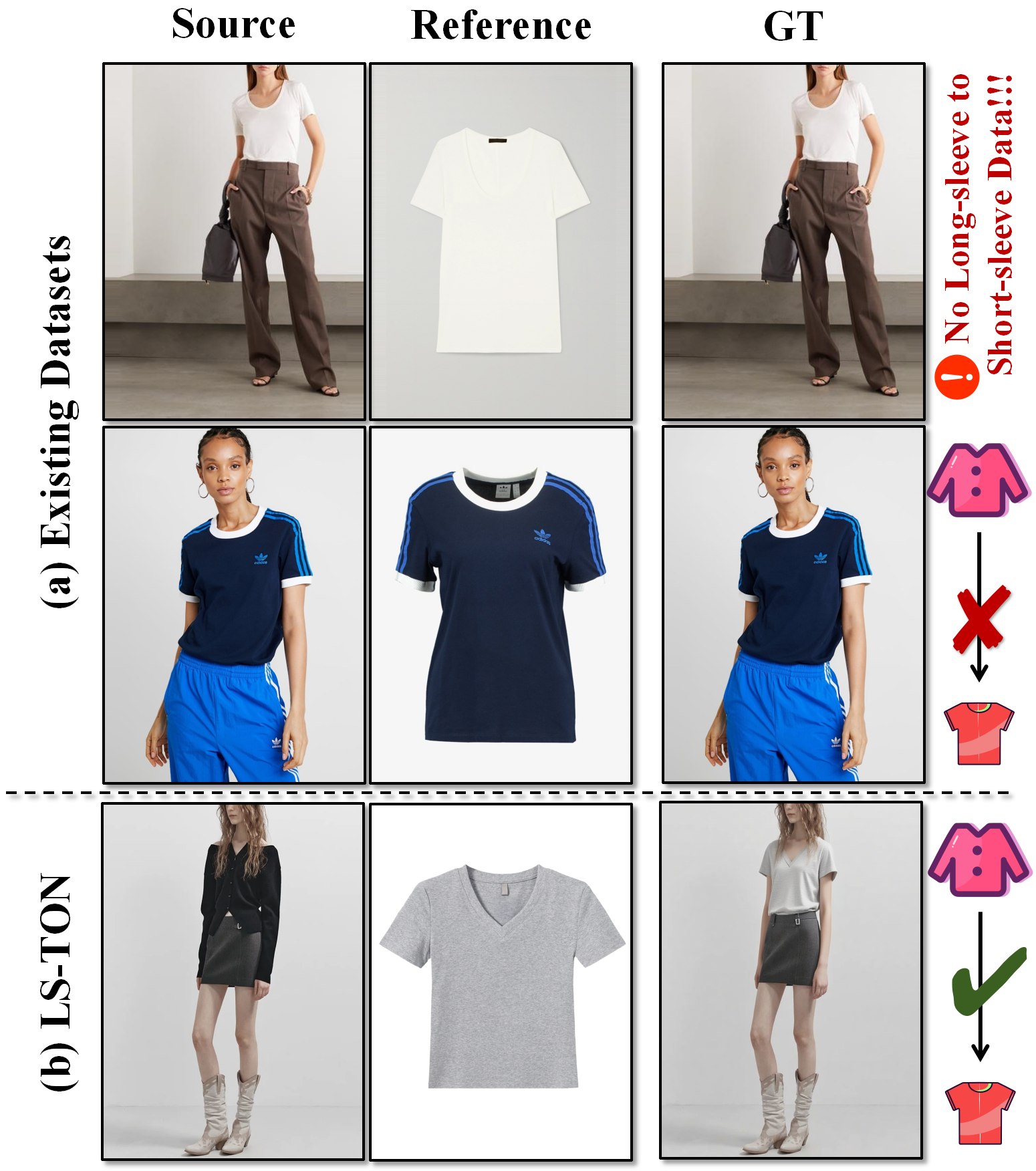}
    \caption{The overview of virtual try-on datasets. (a) Existing datasets such as VITON-HD and DressCode lack data for long-sleeve to short-sleeve transformations. (b) LS-TON is a dataset specifically constructed for long-sleeve to short-sleeve garment virtual try-on.}
    \label{dataset}
\end{figure}

\section{Related Work}
\subsection{Image-based Virtual Try-On}
Image-based virtual try-on aims to generate images of a person wearing a target garment while retaining texture details and identity. Typically, methods consist of two stages: garment warping and fusion. Early methods used Thin-Plate Spline (TPS) \cite{bookstein2002principal} for geometric deformation \cite{ge2021disentangled, yang2020towards, han2018viton}, whereas recent optical flow approaches \cite{bai2022single, ge2021parser, he2022style} improve non-rigid alignment but remain prone to estimation errors. Fusion techniques can
be divided into GAN- and diffusion-based approaches. GAN-based methods \cite{lee2022high, lewis2021tryongan, morelli2022dress} demonstrate strong potential but suffer from unstable training \cite{gulrajani2017improved} and mode collapse \cite{miyato2018spectral}, harming image quality. Diffusion-based methods \cite{kim2024stableviton, zhu2023tryondiffusion, yang2024texture} provide more stable training and superior quality with enhanced coverage and flexibility. TryOnDiffusion \cite{zhu2023tryondiffusion} uses dual UNets \cite{ronneberger2015u} to integrate clothing, body, and pose. LaDI-VTON \cite{morelli2023ladi} aligns garment features with CLIP \cite{radford2021learning} embeddings and refines diffusion via warped inputs. GP-VTON \cite{xie2023gp} applies part-wise warping to maintain semantic consistency and texture fidelity. OOTDiffusion \cite{xu2025ootdiffusion} fine-tunes a pre-trained garment UNet within the denoising UNet for detailed learning. IDM-VTON \cite{choi2024improving} merges IP-Adapter \cite{ye2023ip} semantics through cross-attention and employs parallel UNets for refinement. CatVTON \cite{chong2024catvton} simplifies UNet via spatial concatenation, achieving high quality with 899.06 M parameters (49.57 M trainable). Leffa \cite{zhou2025learning} embeds flow-field learning regularization into the diffusion model’s attention layers to align target queries with reference-image regions. This reduces fine-grained distortions and preserves high-quality results in virtual try-on.


Despite these advances, existing methods struggle with long-sleeve to short-sleeve transformations. We propose a simple, effective, training-free universal framework to address this challenge via task decomposition.

\subsection{Classifier‐Free Guidance and Structural Refiner}

Classifier-Free Guidance (CFG) \cite{ho2022classifier} enhances sample quality and controllability in diffusion models and underpins systems like Stable Diffusion \cite{rombach2022high}, DALL·E 2 \cite{ramesh2022hierarchical}, and Imagen \cite{saharia2022photorealistic}. CFG is trained by randomly omitting conditioning signals, enabling the model to predict both conditional and unconditional noise. At inference, these predictions are combined as:
\begin{gather}
    \varepsilon _{\text{uncond}} = \varepsilon_\theta(z_t) , \quad \varepsilon _{\text{cond}} = \varepsilon_\theta(z_t|y) , \\
    \varepsilon _{\text{cfg}}= \varepsilon _{\text{uncond}} + 
\omega (\varepsilon _{\text{cond}}- \varepsilon _{\text{uncond}}), 
\end{gather}
where $\omega$ denotes the guidance scale, $z_t$ represents the latent vector at time step $t$. $\varepsilon _{\text{uncond}}$ and $\varepsilon _{\text{cond}}$ respectively refer to the noise predicted by the model’s forward diffusion process under an empty prompt and under the conditioning prompt $y$. This mechanism balances fidelity and diversity without an external classifier. Virtual try-on methods \cite{choi2024improving, chong2024catvton, zhou2025learning} typically use CFG with $\omega = 2.5$ to maintain texture fidelity and realism, but a fixed scale may not suit specialized tasks \cite{shen2024rethinking, li2025adaptive}. We thus propose Dynamic CFG to adaptively adjust $\omega$ during sampling.

Recent studies integrate frequency-domain knowledge into image enhancement to better restore details and structure \cite{chen2024large, ma2024learning}. For example, Fuoli et al. \cite{fuoli2021fourier} introduce a Fourier-domain perceptual loss to maintain high-frequency consistency and sharpen edges and textures in super-resolved images. Zhou et al. \cite{zhou2022spatial} propose a spatial–frequency fusion network that jointly extracts and fuses spatial and frequency features to enhance global and local representations, significantly improving pan-sharpening performance. Building on these works, we leverage structural cues from the input image’s frequency domain to further improve output quality.

\section{LS-TON Benchmark}
The widely used public virtual try-on datasets, such as VITON-HD \cite{choi2021viton} and DressCode \cite{morelli2022dress}, lack data specifically designed for the long-sleeve to short-sleeve garment transfer task, most instances involve short-sleeve to short-sleeve transformations, as depicted in Fig. \ref{dataset}(a). Consequently, these datasets are insufficient for effectively evaluating the performance of VTON models on the more challenging long-sleeve to short-sleeve scenario. To bridge this gap, we introduce a novel benchmark, LS-TON, tailored for long-sleeve to short-sleeve transfer. We curated image pairs from social platforms, consisting of models dressed in long-sleeved garments (e.g., heavy coats, padded jackets, and long-sleeved shirts) and corresponding short-sleeved clothing. Using a variety of image editing tools, we manually replace the long sleeves on the models with short sleeves, resulting in image triplets: the source image (the model in long sleeves), the reference image (the short-sleeved garment), and the ground-truth image (the model in short sleeves), as shown in Fig. \ref{dataset}(b). LS-TON comprises 770 such triplets.

\section{Method}

We aim to achieve seamless and realistic virtual try-on of short-sleeved garments for models originally dressed in long-sleeved clothing. To this end, we introduce UR-VTON, a training-free framework capable of integrating any VTON model. As illustrated in Fig. \ref{UR-VTON}, UR-VTON comprises three key components: an ``Undress-to-Redress'' module, Dynamic Classifier-Free Guidance (DCFG) for diffusion process, and Structural Refiner to enhance output quality. Next, we present the detailed description.

\subsection{Undress-to-Redress}
Existing VTON models face a practical challenge in transforming long sleeves to short sleeves, owing to the ``majority'' rule: when most pixels in the target region correspond to skin, the model accurately reconstructs skin; however, when clothing pixels predominate, it often misrepresents skin areas as clothing textures (see Fig. \ref{motivation}). This issue limits VTON’s applicability in general scenarios and undermines the precision of long-to-short sleeve conversions.

To address this, we propose an ``Undress-to-Redress'' mechanism, which decomposes the long-sleeve to short-sleeve transformation into two sequential stages (see Fig. \ref{UR-VTON} (a)): ``intermediate body exposure'' and ``final garment overlay''. In the first stage, an intermediate result is generated with greater torso visibility. Formulation is as follows:
\begin{gather}
    \varepsilon^1 _\theta = \text{UNet}([z^1_t, G^1, M, D], t), \\
    \varepsilon^1 _{\text{cond}}, \varepsilon^1 _{\text{uncond}} = \text{Chunk}(\varepsilon^1 _\theta), \\
    \varepsilon^1 _{\text{cfg}} = \varepsilon^1 _{\text{uncond}} + \omega (\varepsilon^1 _{\text{cond}}- \varepsilon^1 _{\text{uncond}}), \\
    z^1_{t-1} = \frac{1}{\sqrt{\alpha _t} }(z^1_t - \frac{1-\alpha _t}{\sqrt{1-\bar{\alpha} _t} }\varepsilon^1 _{cfg}) + \sigma _t\varepsilon'  , 
\end{gather}
where $z^1_t$ is the latent representation from forward diffusion. $G^1$ is the reference image for replacement (we choose the bra). $M$ and $D$ denote the binary mask of source image and DensePose, respectively. $t$ is the time step. $\alpha_t$ and $\bar{\alpha}_t$ are forward diffusion retention coefficients. $\sigma_t$ is the noise standard deviation from the scheduler. $\varepsilon' \sim \mathcal{N}(0, I)$. $\text{Chunk}$ represents the average operation on the batch dimension. The guidance scale $\omega$ is often empirically set to 2.5.

Similarly, in the next stage, the user-provided reference image of the short-sleeved garment $G^2$ is virtually applied to the intermediate image. The formulation is as follows:
\begin{gather}
    \varepsilon^2 _\theta = \text{UNet}([z^2_t, G^2, M, D], t), \\
    \varepsilon^2 _{\text{cond}}, \varepsilon^2 _{\text{uncond}} = \text{Chunk}(\varepsilon^2 _\theta),  
\end{gather}
\begin{gather}
    \varepsilon^2 _{\text{cfg}} = \varepsilon^2 _{\text{uncond}} + \omega (\varepsilon^2 _{\text{cond}}- \varepsilon^2 _{\text{uncond}}), \\
    z^2_{t-1} = \frac{1}{\sqrt{\alpha _t} }(z^2_t - \frac{1-\alpha _t}{\sqrt{1-\bar{\alpha} _t} }\varepsilon^2 _{cfg}) + \sigma _t\varepsilon' ,   
\end{gather}
where $z^2_t$ is the latent representation from forward diffusion on the intermediate image.

The rationale for this two-stage approach is that direct transformation from long sleeves to short sleeves is often intractable. By first generating an image with the model dressed only in undergarments, the arms and shoulders become clearly visible, thus alleviating the need for the model to simultaneously render both the torso and garment regions. 

\subsection{Dynamic Classifier-Free Guidance}
Classifier-Free Guidance (CFG) \cite{ho2022classifier} is a technique for adjusting the intensity of conditional signals in diffusion models without requiring an additional classifier. At each denoising step, CFG linearly combines noise predictions from conditional and unconditional models, enabling a controllable trade-off between sample fidelity and diversity. However, existing VTON models typically use a fixed weighting coefficient throughout all denoising steps. Although this ensures stable guidance, it frequently results in excessive guidance at early stages and insufficient guidance at later stages, introducing unwanted artifacts \cite{kynkaanniemi2024applying}. To address this issue, we propose Dynamic CFG, in Fig. \ref{UR-VTON} (b), inspired by \cite{malarz2025classifier}, which employs dynamic weighting coefficients to adaptively control the conditional signal strength. This approach preserves diversity during the initial stages and enhances output quality at later stages. Taking the first stage as an example, the formulation is presented as follows:
\begin{gather}
    \omega_{\text{dcfg}} = \omega -\cos (\pi\times \frac{t}{T-1} ), \\
    \varepsilon^1 _{\text{cfg}}= \varepsilon^1 _{\text{uncond}} + \omega_{\text{dcfg}} (\varepsilon^1 _{\text{cond}}- \varepsilon^1 _{\text{uncond}}),
\end{gather}
where $t$ represents the current timestep, and $T$ indicates the total number of timesteps.

\subsection{Structural Refiner}
To further enhance the edge structures in the virtual try-on results, in Fig. \ref{UR-VTON} (c), we incorporate the structural information of the input image $P(x,y)$ into the generated output $R(x,y)$, thus improving the quality of the synthesis. Specifically, we first apply the Fourier transform to convert both $P(x,y)$ and $R(x,y)$ into the frequency domain:
\begin{gather}
    \mathcal{K}_P(u, v) = \mathcal{F}\left \{ P(x,y) \right \} (u, v), \\
    \mathcal{K}_R(u, v) = \mathcal{F}\left \{ R(x,y) \right \} (u, v),
\end{gather}
where $(x, y)$ denotes the spatial coordinates of image pixels, while $(u, v)$ represents the corresponding frequency components. $\mathcal{F}$ refers to the two-dimensional discrete Fourier transform. 

Subsequently, the magnitude and phase components are separated:
\begin{gather}
    \mathcal{A}_P(u, v) = \left | \mathcal{K}_P(u, v) \right | , \\
     \mathcal{A}_R(u, v) = \left | \mathcal{K}_R(u, v) \right |, \quad \Phi _R(u, v) = \angle \mathcal{K}_R(u, v). 
\end{gather}

We apply a high-frequency mask $\mathcal{B}(u, v)$ to fuse the amplitude $\mathcal{A}_P(u, v)$ into $\mathcal{A}_R(u, v)$ in the high-frequency regions. The resulting amplitude is then combined with the phase spectrum $\Phi _R(u, v)$ to form a complex frequency-domain representation, as formulated below:
\begin{align}
\mathcal{A}_{\hat{R}}(u, v) &= [1- \mathcal{B}(u, v)] \mathcal{A}_R(u, v) \notag\\
    &\quad + \frac{\mathcal{B}(u, v)\left[\mathcal{A}_P(u, v)+\mathcal{A}_R(u, v)\right]}{2}, \\
\mathcal{K}_{\hat{R}}(u, v) &= \mathcal{A}_{\hat{R}}(u, v)\exp\left(j \Phi _R(u, v)\right),
\end{align}
where $j$ denotes the imaginary unit.

This composite frequency representation is transformed back into the spatial domain using the inverse Fourier transform, producing an image with enhanced edges $\hat{R}(x, y) = \mathcal{F}^{-1}\left \{ \mathcal{K}_{\hat{R}}(u, v) \right \} (x,y)$.

\section{Experiments}
\subsection{Experimental Settings}
\noindent \textbf{Datasets.} 
We conduct experiments on two public fashion datasets, VITON-HD \cite{choi2021viton} and DressCode \cite{morelli2022dress}, and a custom benchmark (LS-TON) for long-sleeve to short-sleeve try-on evaluation. VITON-HD contains 13,679 image pairs (11,647 training; 2,032 testing), each comprising a front-view upper-body photograph and its corresponding in-shop top at 1024$\times$768 resolution. DressCode contains 48,392 training and 5,400 testing pairs of full-body, front-view images paired with in-shop garments (tops, bottoms, and dresses), also at 1024$\times$768 resolution. LS-TON includes 770 triplets: a person wearing a long-sleeved garment, the matching in-shop short-sleeved garment, and the person wearing the short-sleeved version. The long-sleeved garments span thick coats, hoodies, and padded jackets. All LS-TON data follow the VITON-HD preprocessing protocol.

\noindent \textbf{Baselines.} We benchmark our proposed UR-VTON against several state-of-the-art diffusion-based virtual try-on models, including:  GP-VTON \cite{xie2023gp} (part-wise garment warping for semantic and texture fidelity), LaDI-VTON \cite{morelli2023ladi} (CLIP-based feature alignment and diffusion refinement), IDM-VTON \cite{choi2024improving} (cross-attention and parallel UNets for detail), OOTDiffusion \cite{xu2025ootdiffusion} (fine-tuned garment UNet for denoising), CatVTON \cite{chong2024catvton} (UNet simplification via spatial concatenation), and Leffa \cite{zhou2025learning} (attention flow regularization for artifact reduction and efficient portrait generation).

\noindent \textbf{Evaluation Metrics.} 
To ensure fair evaluation of our model, we follow the established VTON evaluation protocol. In the paired try-on setting with ground truth from the test dataset, we employ four widely used metrics to assess the similarity between synthesized and real images: Structural Similarity Index (SSIM) \cite{wang2004image}, Learned Perceptual Image Patch Similarity (LPIPS) \cite{zhang2018unreasonable}, Fréchet Inception Distance (FID) \cite{fid}, and Kernel Inception Distance (KID) \cite{binkowski2018demystifying}. In the unpaired setting, we use FID and KID to evaluate the distributional similarity between generated and real samples.

\noindent \textbf{Implementation Details.} 
The experiments are conducted on an Ubuntu 20.04.6 LTS system with the PyTorch framework, a single NVIDIA L20 GPU, and 45 GB of RAM. Unless otherwise specified, Leffa is used as the backbone for UR-VTON in all subsequent experimental settings. The total number of timesteps $T=30$, and guidance scale $\omega = 2.5$.

\begin{table}[t]
\centering
\resizebox{0.47\textwidth}{!}{
\begin{tabular}{@{}l|c|cccccc@{}}
   \toprule
    \multicolumn{2}{c|}{\multirow{2}{*}{Methods}} & \multicolumn{6}{c}{LS-TON} \\
    \cmidrule(r){3-8}
     &  & \multicolumn{4}{c}{Paired} & \multicolumn{2}{c}{Unpaired} \\
    \cmidrule(r){1-2} \cmidrule(r){3-6} \cmidrule(r){7-8}
     Models & Backbone & SSIM $\uparrow$ & FID $\downarrow$ & KID $\downarrow$ & LPIPS $\downarrow$ & FID $\downarrow$ & KID $\downarrow$ \\
     \midrule
     LaDI-VTON [2023 ACM MM] & - & 0.768 & 47.848 & 23.863 & 0.206 & 50.593 & 26.120 \\
     IDM-VTON [2024 ECCV] & - & 0.750 & 43.311 & 14.752 & 0.247 & 46.601 & 19.923 \\
     OOTDiffusion [2025 AAAI] & - & 0.775 & 50.615 & 24.762 & 0.199 & 50.775 & 23.460 \\
     CatVTON [2025 ICLR] & - & 0.760 & 40.702 & 11.736 & 0.227 & 41.401 & 11.866
     \\
     Leffa [2025 CVPR] & - & 0.755 & 41.398 & 13.405 & 0.234 & 45.557 & 14.001 \\
     \hline
     \hline
     \textbf{UR-VTON} & CatVTON & \underline{0.804} & \underline{31.315} & \underline{8.038} & \textbf{0.154} & \underline{33.094} & \textbf{9.090} \\ 
     
     \textbf{UR-VTON} & Leffa & \textbf{0.811} & \textbf{31.074} & \textbf{7.723} & \underline{0.169} & \textbf{32.866} & \underline{8.322} \\
    
   \bottomrule
\end{tabular}}
\caption{Quantitative comparison on LS-TON. The best and the second best-results are \textbf{highlighted} and \underline{underlined}.}
\label{lston}
\end{table}

\begin{table*}[htb!]
\centering
\resizebox{1\textwidth}{!}{
\begin{tabular}{@{}l|c|cccccc|cccccc@{}}
   \toprule
    \multicolumn{2}{c|}{\multirow{2}{*}{Methods}} & \multicolumn{6}{c|}{VITON-HD} & \multicolumn{6}{c}{DressCode} \\
    \cmidrule(r){3-8} \cmidrule(r){9-14}
     &  & \multicolumn{4}{c}{Paired} & \multicolumn{2}{c|}{Unpaired} & \multicolumn{4}{c}{Paired} & \multicolumn{2}{c}{Unpaired}  \\
    \cmidrule(r){1-2} \cmidrule(r){3-6} \cmidrule(r){7-8} \cmidrule(r){9-12} \cmidrule(r){13-14}
     Models & Backbone & SSIM $\uparrow$ & FID $\downarrow$ & KID $\downarrow$ & LPIPS $\downarrow$ & FID $\downarrow$ & KID $\downarrow$ & SSIM $\uparrow$ & FID $\downarrow$ & KID $\downarrow$ & LPIPS $\downarrow$ & FID $\downarrow$ & KID $\downarrow$ \\
     \midrule
     GP-VTON [2023 CVPR] & - & \underline{0.891} & 6.539 & 1.097 & 0.069 & 9.628 & 1.298 & \textbf{0.922} & 7.466 & 5.738 & 0.070 & 10.064 & 8.035 \\
     LaDI-VTON [2023 ACM MM] & - & 0.864 & 11.211 & 7.006 & 0.073 & 14.418 & 8.422 & 0.895 & 7.280 & 4.479 & 0.057 & 9.402 & 5.421 \\
     IDM-VTON [2024 ECCV] & - & \textbf{0.899} & 5.762 & 0.732 & 0.079 & 9.842 & 2.134 & \underline{0.904} & 5.165 & 2.553 & 0.072 & 6.941 & 2.979 \\
     OOTDiffusion [2025 AAAI] & - & 0.842 & 6.236 & 0.770 & 0.086 & 9.341 & 1.018 & 0.886 & 5.341 & 2.372 & 0.090 & 8.929 & 4.399 \\
     CatVTON [2025 ICLR] & - & 0.835 & 6.389 & 0.894 & 0.070 & 9.050 & 1.181 & 0.877 & \underline{4.049} & \underline{0.981} & \underline{0.053} & \underline{6.137} & \underline{1.403}
     \\
     Leffa [2025 CVPR] & - & 0.8566 & \textbf{5.261} & \textbf{0.154} & \underline{0.063} & \textbf{8.357} & \textbf{0.227} & 0.893 & 5.130 & 1.941 & 0.107 & 6.370 & 2.309 \\
     \hline
     \hline
     \textbf{UR-VTON} & CatVTON & 0.869 & \underline{5.676} & \underline{0.547} & \textbf{0.060} & 9.275 & 1.496 & 0.902 & \textbf{3.214} & \textbf{0.644} & \textbf{0.042} & \textbf{5.207} & \textbf{1.300}  \\ 
     
     \textbf{UR-VTON} & Leffa & 0.844 & 6.218 & \underline{0.531} & 0.077 & \underline{9.028} & \underline{0.689} & 0.883 & 5.334 & 2.188 & 0.119 & 7.001 & 2.593 \\
    
   \bottomrule
\end{tabular}}
\caption{Quantitative comparison on VITON-HD and DressCode. The best and the second-best results are \textbf{highlighted} and \underline{underlined}.}
\label{exp1}
\end{table*}

\subsection{Quantitative Comparison}
To comprehensively evaluate the performance of UR-VTON, we conduct quantitative comparisons with state-of-the-art models on the VITON-HD, DressCode, and our proprietary LS-TON datasets. Experiments are performed under both paired and unpaired settings to assess not only the similarity between synthesized images and ground truth, but also the models’ generalization capabilities. Notably, as UR-VTON can integrate arbitrary VTON backbones, we implemented both CatVTON and Leffa as its backbone in this experiment settings. Table \ref{lston} shows that UR-VTON significantly outperforms all baseline methods across every metric on LS-TON, demonstrating the effectiveness of the UR-VTON model architecture and the ``undress-to-redress'' strategy in long-sleeve to short-sleeve conversions. Additionally, we evaluate UR-VTON on VITON-HD and DressCode to observe its ability in general scenarios. As reported in Table \ref{exp1}, UR-VTON still achieves the best or second-best results on most metrics, further confirming its superiority. Notably, Leffa achieves competitive results on VITON-HD, especially in terms of the KID metric. We attribute this to the nature of the KID score, which predominantly measures distributional differences between generated outputs and ground truth. Since UR-VTON requires additional torso reconstruction during synthesis, the distribution calculations over the generated torso area may lead to higher KID values. In contrast, Leffa focuses solely on the garment region and does not involve torso reconstruction.

\begin{figure*}[htb!]
    \centering
    \includegraphics[width=0.98\linewidth]{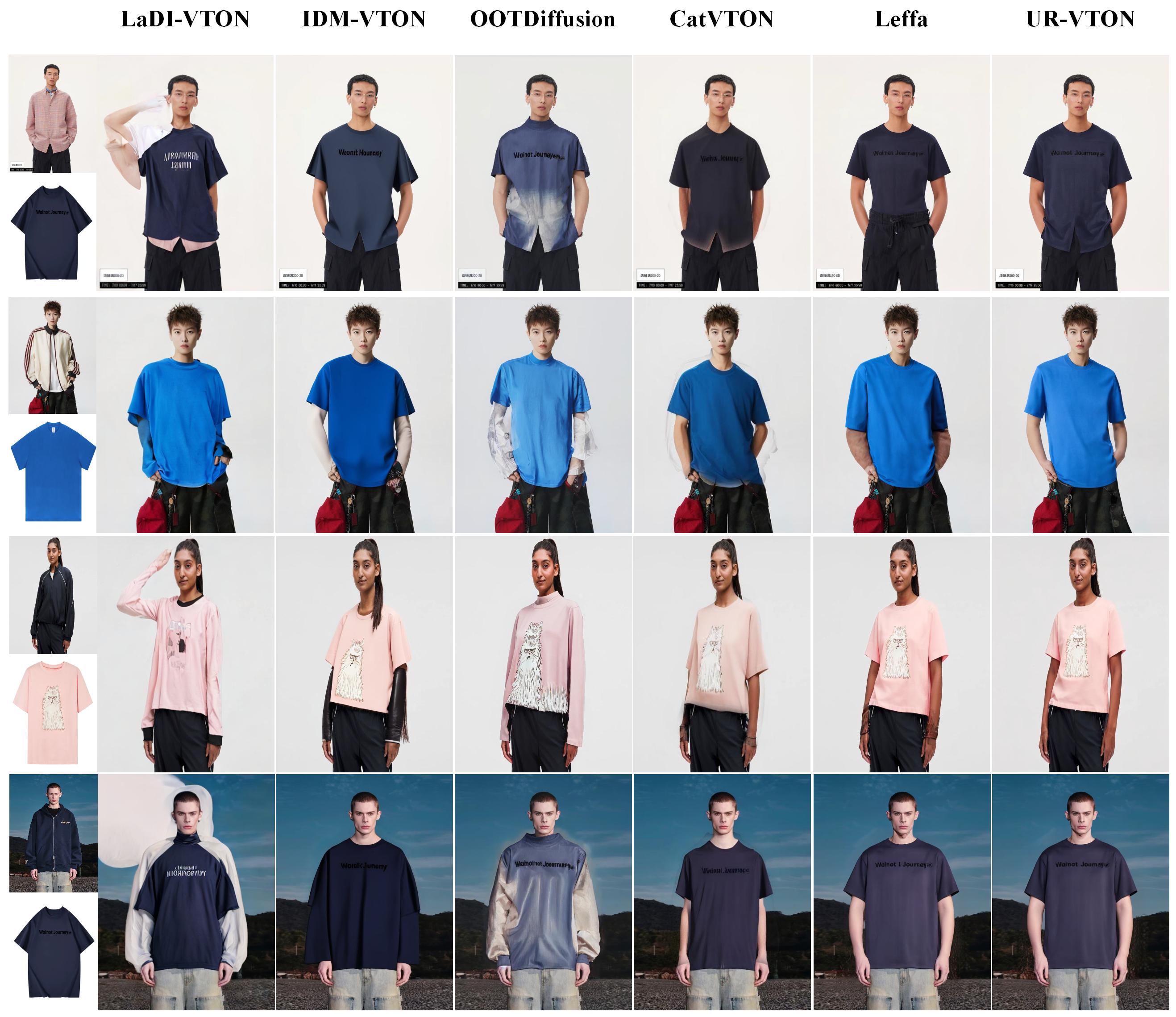}
    \caption{\textbf{Qualitative comparison on the LS-TON dataset.} UR-VTON demonstrates significant advantages when handling the challenging task of converting long-sleeve to short-sleeve garment transition scenarios. Please zoom in to view more details.}
    \label{visual}
\end{figure*}

\begin{table}[t]
\centering
\resizebox{0.47\textwidth}{!}{
\begin{tabular}{@{}lcccccc@{}}
   \toprule
    \multirow{2}{*}{Methods} & \multicolumn{4}{c}{Paired} & \multicolumn{2}{c}{Unpaired} \\
    \cmidrule(r){2-5} \cmidrule(r){6-7}
     & SSIM $\uparrow$ & FID $\downarrow$ & KID $\downarrow$ & LPIPS $\downarrow$ & FID $\downarrow$ & KID $\downarrow$ \\
     \midrule
     Original & 0.755 & 41.398 & 13.405 & 0.234 & 45.557 & 14.001 \\
     + UR & 0.795 & 33.538 & 8.796 & 0.186 & 34.487 & 9.896 \\
     + DCFG & 0.762 & 41.195 & 11.885 & 0.229 & 44.741 & 12.946 \\
     + SR & 0.759 & 41.288 & 12.694 & 0.230 & 45.018 & 13.529 \\
     \textbf{UR-VTON} & \textbf{0.811} & \textbf{31.074} & \textbf{7.723} & \textbf{0.169} & \textbf{32.866} & \textbf{8.322} \\
   \bottomrule
\end{tabular}}
\caption{Ablation study on LS-TON. The best results are \textbf{highlighted}. ``UR'' represents the Undress-to-Redress mechanism. ``DCFG'' represents the dynamic classifier-free guidance. ``SR'' represents the structural refiner.}
\label{abla}
\end{table}

\begin{figure*}[htb!]
    \centering
    \includegraphics[width=0.93\linewidth]{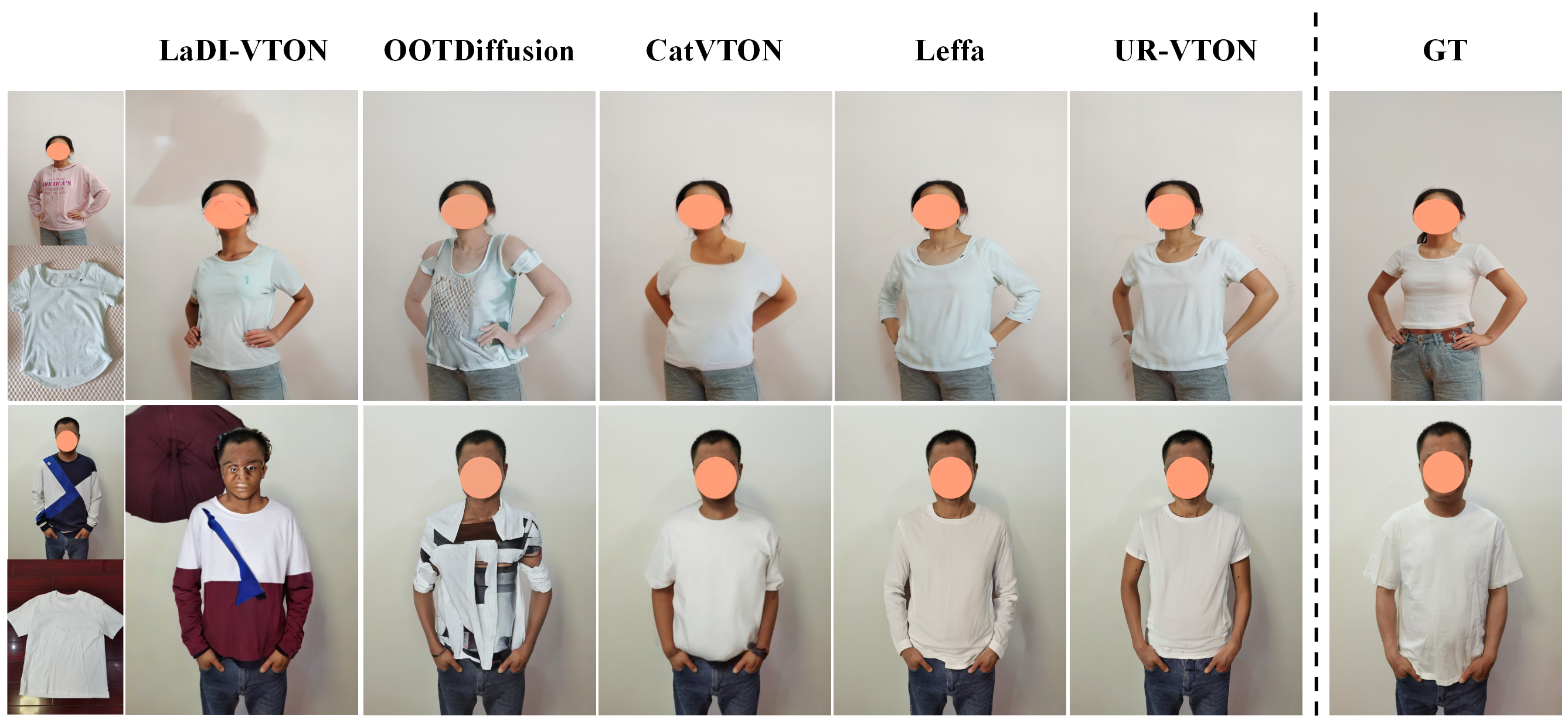}
    \caption{\textbf{Qualitative comparison of UR-VTON in real-world long-sleeve–to–short-sleeve garment transitions.} Please zoom in to view more details. ``GT'' denotes the ground-truth.}
    \label{userstudy}
\end{figure*}

\subsection{Qualitative Comparison}
To further demonstrate the effectiveness of UR-VTON, we conduct a qualitative evaluation of its virtual try-on results via visual analysis in long-sleeve to short-sleeve garment conversion tasks. As illustrated in Fig. \ref{visual}, we showcase challenging sleeve transformation results on the LS-TON dataset and compare multiple state-of-the-arts methods to highlight differences in fine-grained consistency. Notably, alternative approaches often exhibit incorrect garment transfer, visual artifacts, chaotic outputs, and detail loss, which substantially limit their real-world applicability. In contrast, UR-VTON consistently produces accurate and natural garment transfers, further highlighting its superior generalization capability and practical utility.

\subsection{Ablation Study}
We conduct ablation studies on the LS-TON dataset to assess the contribution of each module to UR-VTON’s performance in long-sleeve-to-short-sleeve garment conversion. As shown in Table \ref{abla}, the model employing only the  ``Undress-to-Redress'' (UR) mechanism exhibits a substantial improvement over the original model. In contrast, using dynamic classifier-free guidance (DCFG) or structural refiner (SR) alone results in only modest gains. Ultimately, the UR-VTON model, which integrates UR, DCFG, and SR, achieves the highest performance. This outcome is anticipated, as the UR mechanism offers an intermediate representation with increased torso exposure, significantly reducing task complexity and enabling marked performance enhancements. DCFG and SR primarily serve to refine the fine-grained details of the try-on outputs. The integration of these three modules provides comprehensive and stable global information, thereby offering strong support for virtual try-on systems.

\subsection{Application}

To assess the practical effectiveness of UR-VTON, we recruited several participants as models and photographed them in both long-sleeved and short-sleeved garments using a mobile device, with an emphasis on maintaining consistent poses. The collected images are subsequently processed by the VTON models. As illustrated in Fig. \ref{userstudy}, UR-VTON demonstrated superior vitrual try-on performance, producing highly accurate results. By contrast, alternative methods failed to achieve precise garment transfer, further substantiating the practical advantages of UR-VTON.

\section{Conclusion}
This paper introduces a novel training-free virtual try-on (VTON) framework, UR-VTON, which utilizes an ``Undress-to-Redress'' mechanism alongside two detail refinement strategies to address the challenge of transferring from long-sleeved to short-sleeved garments. By decomposing the process into sequential steps, the proposed method effectively prevents unrealistic image generation resulting from insufficient torso exposure in direct synthesis. Moreover, the integration of fine-grained information mitigates artifacts that may emerge during multiple inference stages. To effectively assess the performance of VTON models for this specific task, we construct a novel benchmark dataset, LS-TON, dedicated to the long-sleeve-to-short-sleeve garment transition. Experimental results demonstrate that UR-VTON consistently outperforms existing methods across diverse datasets and tasks. Additionally, application confirms that UR-VTON is highly practical and exhibits strong applicability in real-world scenarios.



\bibliography{aaai2026}

\appendix
\section{Appendix}
\begin{figure*}[htb!]
    \centering
    \includegraphics[width=0.8\textwidth]{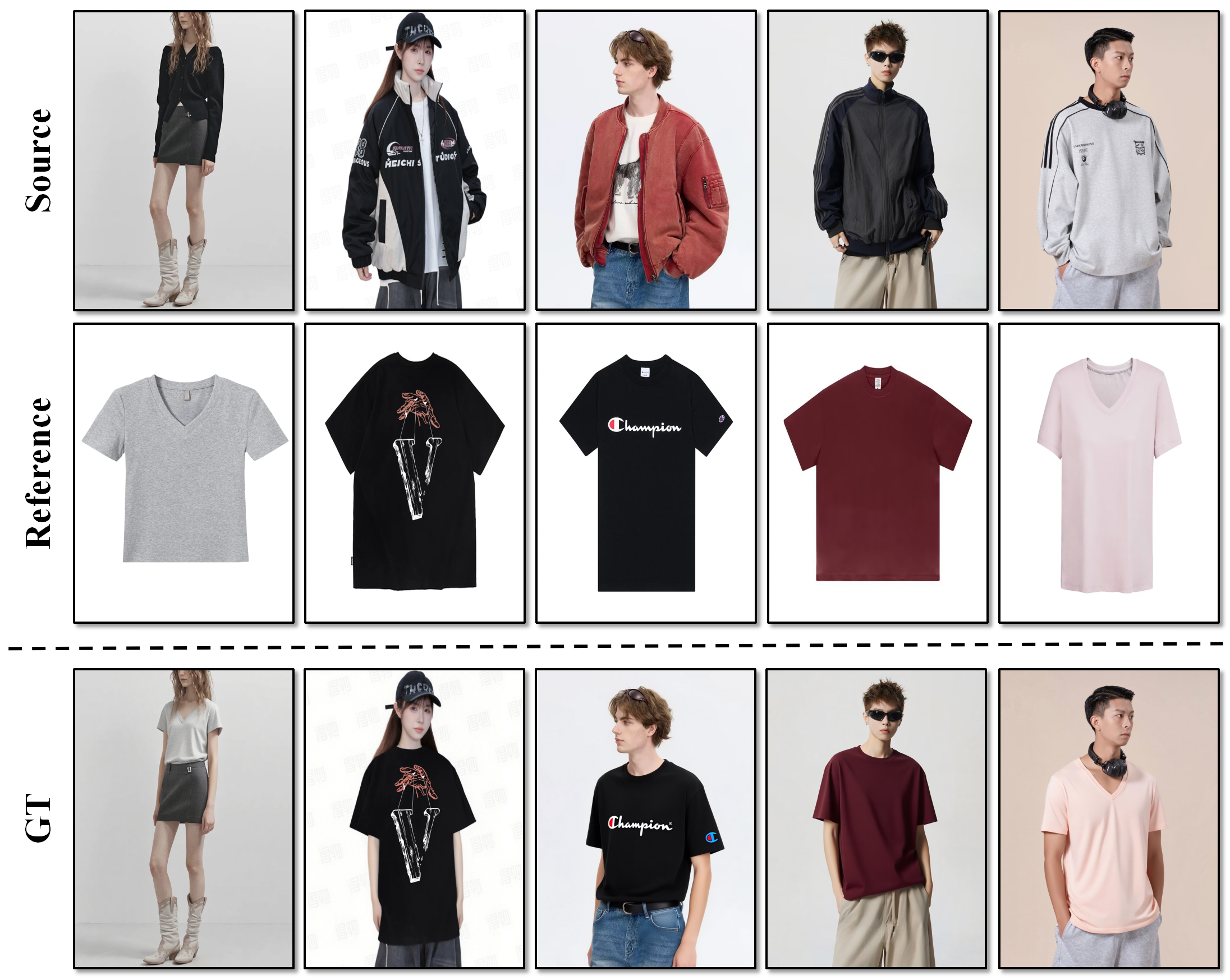}
    \caption{The overview of long-sleeve to short-sleeve try-on dataset.}
    \label{rebuttal_lston}
\end{figure*}

\subsection{LS-TON Benchmark}
Widely used public virtual try‑on datasets—specifically VITON‑HD \cite{choi2021viton} and DressCode \cite{morelli2022dress}—lack data curated for the long‑sleeve‑to‑short‑sleeve conversion task. Consequently, these datasets are inadequate for effectively evaluating VTON models in the more challenging scenario of long‑sleeve to short‑sleeve transformations. To address this gap, we introduce a new benchmark, LS‑TON, specifically designed for long‑sleeve to short‑sleeve conversions. We curated image pairs from social platforms, consisting of models dressed in long-sleeved garments (e.g., heavy coats, padded jackets, and long-sleeved shirts) and short-sleeved clothing. Using multiple image editing tools, we manually replaced the long‑sleeve garment in the model image with the short‑sleeve garment, producing an image triplet: the source image (model wearing long sleeves), the reference image (short‑sleeve garment), and the ground‑truth (GT) image (model actually wearing the short‑sleeve garment), as illustrated in Fig. \ref{rebuttal_lston}. The resulting LS‑TON dataset comprises 770 such triplets.

\subsection{Additional Experiments}
\subsubsection{Experimental Settings}
\noindent \textbf{Datasets.} 
We conduct experiments on two public fashion datasets, VITON-HD \cite{choi2021viton} and DressCode \cite{morelli2022dress}, and a custom benchmark (LS-TON) for long-sleeve to short-sleeve try-on evaluation. VITON-HD contains 13,679 image pairs (11,647 training; 2,032 testing), each comprising a front-view upper-body photograph and its corresponding in-shop top at 1024$\times$768 resolution. DressCode contains 48,392 training and 5,400 testing pairs of full-body, front-view images paired with in-shop garments (tops, bottoms, and dresses), also at 1024$\times$768 resolution.

\noindent \textbf{Baselines.} We benchmark our proposed UR-VTON against several state-of-the-art diffusion-based virtual try-on models, including:  GP-VTON \cite{xie2023gp} (part-wise garment warping for semantic and texture fidelity), LaDI-VTON \cite{morelli2023ladi} (CLIP-based feature alignment and diffusion refinement), IDM-VTON \cite{choi2024improving} (cross-attention and parallel UNets for detail), OOTDiffusion \cite{xu2025ootdiffusion} (fine-tuned garment UNet for denoising), CatVTON \cite{chong2024catvton} (UNet simplification via spatial concatenation), and Leffa \cite{zhou2025learning} (attention flow regularization for artifact reduction and efficient portrait generation).


\noindent \textbf{Implementation Details.} 
The experiments are conducted on an Ubuntu 20.04.6 LTS system with the PyTorch framework, a single NVIDIA L20 GPU, and 45 GB of RAM. Unless otherwise specified, Leffa is used as the backbone for UR-VTON in all subsequent experimental settings. The total number of timesteps $T=30$, and guidance scale $\omega = 2.5$.

\begin{figure*}[t]
    \centering
    \includegraphics[width=1\textwidth]{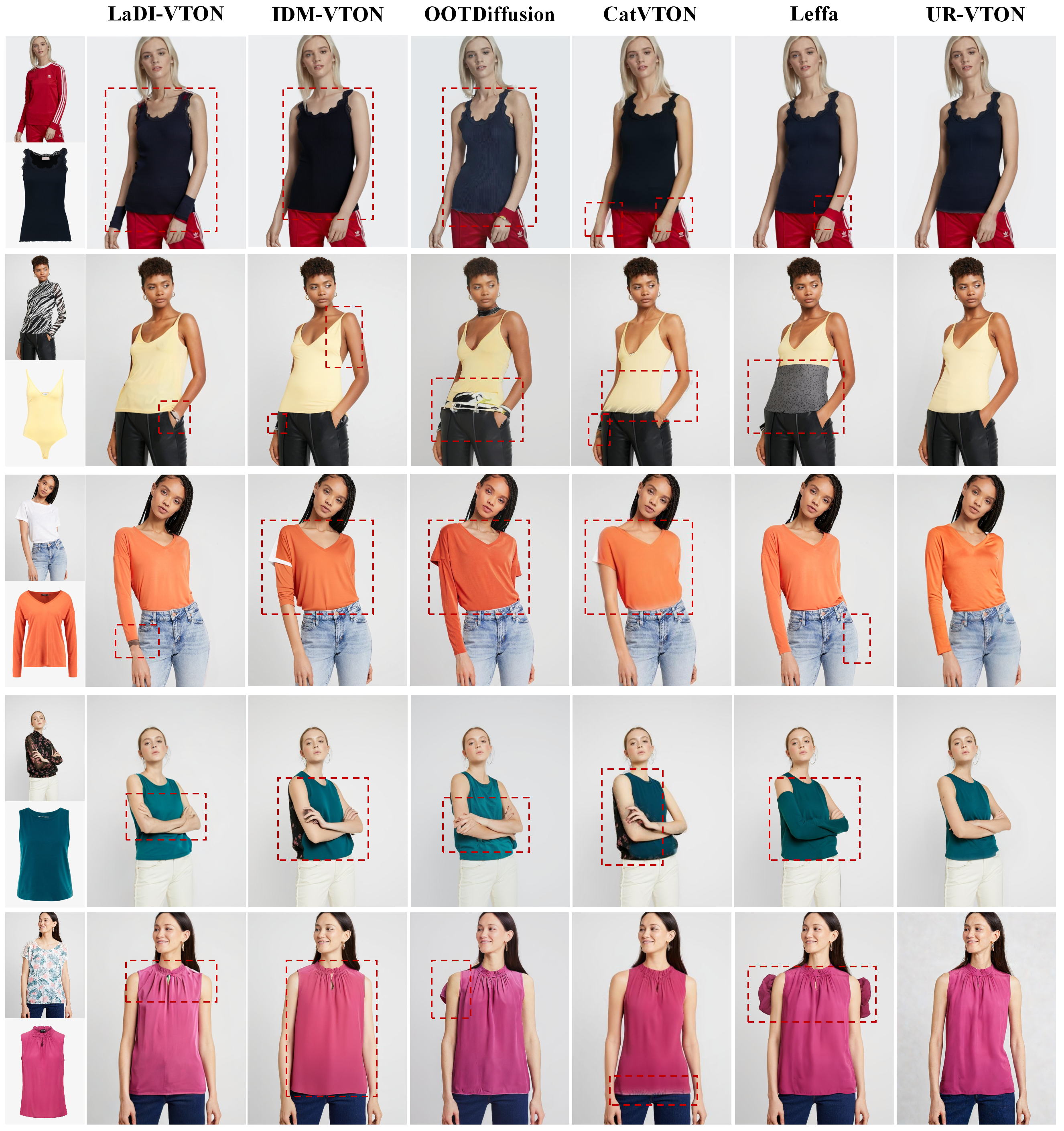}
    \caption{\textbf{Qualitative comparison on the VITON-HD dataset.} UR-VTON demonstrates significant advantages. We highlight the failure regions of the virtual try-on using red bounding boxes. Please zoom in to view more details.}
    \label{rebuttal_HD}
\end{figure*}

\begin{figure*}[t]
    \centering
    \includegraphics[width=1\textwidth]{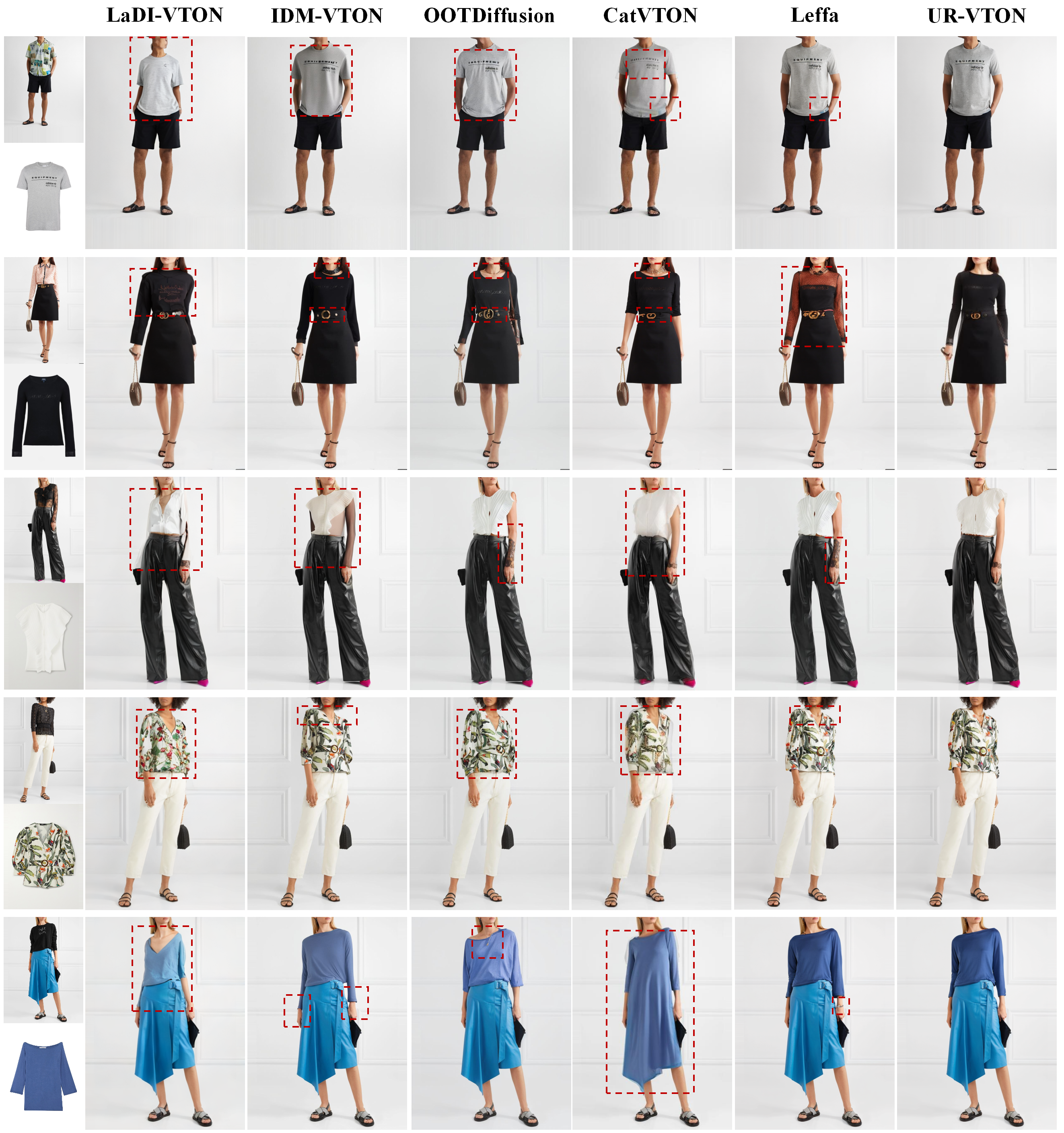}
    \caption{\textbf{Qualitative comparison on the DressCode dataset.} UR-VTON demonstrates significant advantages. We highlight the failure regions of the virtual try-on using red bounding boxes. Please zoom in to view more details.}
    \label{rebuttal_DC}
\end{figure*}

\subsubsection{Qualitative Comparison}
To further demonstrate the effectiveness of UR-VTON, we qualitatively evaluated its performance on arbitrary garment-swap tasks by visually inspecting its virtual try-on results. As shown in Fig. \ref{rebuttal_HD} and \ref{rebuttal_DC}, we test UR-VTON on challenging examples from two widely adopted high-resolution datasets—VITON-HD and DressCode—and compared it with several state-of-the-art methods to highlight differences in fine-grained consistency. Notably, baseline methods frequently exhibit garment transfer errors, visual artifacts, incoherent outputs, and detail loss, significantly limiting their practical applicability. In contrast, UR-VTON consistently produces accurate, natural garment transfers, underscoring its superior generalization and real-world applicability.

\end{document}